\documentclass[10pt,journal]{IEEEtran}
\usepackage[T1]{fontenc}
\usepackage[utf8]{inputenc}
\usepackage{graphicx}
\graphicspath{{./}}
\usepackage{amsmath,amssymb,bm}
\usepackage{booktabs}
\usepackage{tabularx}
\usepackage{array}
\usepackage{url}
\usepackage{balance}
\usepackage[hidelinks,breaklinks]{hyperref}
\begin{document}

\title{SAM3-LoRA: Parameter-Efficient Adaptation of a Concept-Promptable Foundation Model for Multi-Class Structural Defect Segmentation}
\author{P.~Malaisree$^{1,2,3}$, S.~Youwai$^{2}$, S.~Janrungautai$^{1}$, D.~Amorndechaphon$^{3}$, P.~Rojanavasu$^{4}$, and W.~Songkitti$^{5}$\thanks{$^{1}$MAA Consultants Co., Ltd. $^{2}$AI Research Group, Department of Civil Engineering, Faculty of Engineering, King Mongkut\'s University of Technology Thonburi (KMUTT). $^{3}$School of Engineering, University of Phayao. $^{4}$School of Information and Communication Technology, University of Phayao. $^{5}$Thailand Institute of Scientific and Technological Research (TISTR). Corresponding authors: phmq1401@gmail.com; sompote.you@kmutt.ac.th.}}
\maketitle

\begin{abstract}
Promptable segmentation foundation models such as SAM3 accept an open-vocabulary text concept and return every instance matching it, but adapting them to a specialized domain by full fine-tuning is computationally prohibitive for the organizations that would benefit most. This study applies Low-Rank Adaptation (LoRA) to SAM3 for multi-class structural defect segmentation and examines both how such a model can be supervised from conventional annotation and whether the resulting efficiency gain transfers across datasets. Two contributions are methodological. First, we describe a supervision procedure that trains a concept-promptable model directly from COCO-style class-labeled instance segmentation by using the category name itself as the prompt, requiring no prompt templates, no synonym expansion, and no learned class embeddings. Second, we identify and mitigate a failure mode specific to this setting: because a conventional annotation file yields positive prompts exclusively, the model's presence prediction decouples from the text condition and degenerates into responding to any prompt, a collapse that is invisible to every metric computed on positive prompts alone. Exhaustive hard-negative prompting, in which every dataset category absent from an image is issued as a zero-detection query, addresses this at no annotation cost. Two adapter placements were compared under an identical protocol, updating 0.121\% and 1.341\% of model parameters. On a purpose-built tunnel lining dataset, pixel intersection-over-union improved from 0.017 to 0.338 and instance-level recall from 0.375 to 0.672; on the independent public Structural Defects Dataset, from 0.017 to 0.855 and from 0.574 to 1.000. Improvements were directionally consistent across ten metrics on both datasets, and the largest per-category gains occurred precisely where zero-shot competence was absent.
\end{abstract}


\section{Introduction}\label{introduction}

Foundation models pretrained on broad visual corpora have substantially advanced zero-shot and few-shot transfer across computer vision tasks. The Segment Anything Model (SAM) \cite{ref1} established promptable, class-agnostic segmentation as a general-purpose capability; SAM 2 \cite{ref2} extended it to video, and SAM3 \cite{ref3} generalized prompting from geometric primitives to open-vocabulary visual concepts, so that a short noun phrase retrieves every instance of that concept in an image. This shift is consequential for applied segmentation: the interface is no longer a label set fixed at training time, but language.

It also creates a supervision problem that has received little attention. Applied datasets are not distributed as concept prompts. They are distributed as class-labeled instance segmentation, typically in COCO format, in which each annotated object carries an integer category identifier. A conventional segmentation network consumes that identifier directly through a fixed classification head. A concept-promptable model has no such head; it conditions on language. Adapting one to a specialized domain therefore requires an explicit procedure for turning class labels into prompts, and, as we show, requires supervision that a class-labeled dataset does not contain at all.

The computational dimension of the problem is more familiar. The large Vision Transformer backbones of these models make full fine-tuning expensive, typically requiring multi-accelerator clusters. Low-Rank Adaptation (LoRA) \cite{ref4} and related parameter-efficient methods \cite{ref5,ref6} restrict weight updates to low-rank decompositions inserted into an otherwise frozen backbone, and have been applied to SAM in specialized domains, though almost exclusively to single-class crack segmentation built on the first-generation model \cite{ref7,ref8}. Two questions remain open in that literature: whether adapter placement, as distinct from adapter rank, materially changes the outcome, and whether a reported efficiency gain reflects the adaptation method or the idiosyncrasies of the single dataset on which it was measured.

We study these questions in the setting of tunnel lining inspection, which supplies a genuinely multi-class problem (crack, concrete spalling, water ingress) and a deployment context in which the computational constraint is real rather than hypothetical: inspection agencies operate single-GPU workstations, not clusters. We additionally evaluate on the independently collected, publicly available Structural Defects Dataset (S2DS) \cite{ref9} under an identical protocol, so that the efficiency claim is tested rather than assumed.

The contributions of this study are as follows.

\begin{enumerate}
\def\labelenumi{\arabic{enumi}.}
\item
  \textbf{Class-to-concept supervision.} A procedure for training a concept-promptable segmentation model directly from COCO-style class-labeled instance segmentation, in which the category name is used verbatim as the prompt and per-image annotations are grouped into exhaustive per-concept queries (Section~\ref{supervising-a-concept-promptable-model-from-coco-class-labels}).
\item
  \textbf{Hard-negative prompt supervision.} Identification of a presence--text decoupling failure that arises when an open-vocabulary detector is fine-tuned on positive-only prompts, and a two-tier negative prompting scheme that mitigates it at no annotation cost, including its generalization properties and its limits (Section~\ref{hard-negative-prompt-supervision}).
\item
  \textbf{Controlled adapter-placement ablation.} A comparison of two adapter placements at fixed rank, with exact module- and parameter-level accounting: 124 modules updating 0.121\% of parameters against 314 modules updating 1.341\% (Section~\ref{low-rank-adaptation-and-adapter-placement}).
\item
  \textbf{Cross-dataset validation.} Application of the identical baseline, Light, and Full protocol to an independent public benchmark, together with an analysis of where per-category benefit is concentrated and why instance-level metrics understate performance on thin structures (Sections~\ref{results} and~\ref{discussion}).
\end{enumerate}

The remainder of the paper is organized as follows. Section~\ref{related-work} reviews related work. Section~\ref{method} develops the adaptation method. Section~\ref{experimental-setup} describes the experimental setup. Section~\ref{results} reports results, Section~\ref{discussion} discusses their interpretation and limitations, and Section~\ref{conclusion} concludes.

\section{Related Work}\label{related-work}

\subsection{Promptable segmentation foundation models}\label{promptable-segmentation-foundation-models}

SAM \cite{ref1} introduced promptable segmentation trained through a data engine of over one billion masks, achieving zero-shot generalization to novel visual domains without task-specific fine-tuning. SAM 2 \cite{ref2} extended the framework to video via a streaming memory mechanism. SAM3 \cite{ref3} generalized the prompt interface to open-vocabulary concepts, so that a noun phrase retrieves all matching instances rather than a single mask at a point or box. The pretraining distribution of these models is deliberately broad, whereas structural defect segmentation demands accurate delineation of a narrow, site-specific set of categories, which motivates targeted adaptation rather than reliance on pretrained weights alone.

\subsection{Parameter-efficient fine-tuning}\label{parameter-efficient-fine-tuning}

Full fine-tuning updates every parameter and is correspondingly expensive in memory and compute. Parameter-efficient methods restrict trainable capacity to a small subset. Houlsby et al.~\cite{ref5} introduced bottleneck adapter modules inserted between transformer layers. Hu et al.~\cite{ref4} proposed LoRA, which decomposes the weight update into a product of low-rank matrices while the pretrained backbone remains frozen, matching full fine-tuning on many downstream tasks at a fraction of the trainable parameter count. Hayou et al.~\cite{ref6} refined the learning-rate treatment of the two low-rank factors. Collectively these results establish that large pretrained models can be specialized by training a small fraction of their parameters, which forms the methodological basis of the present study.

\subsection{Parameter-efficient adaptation of SAM in specialized domains}\label{parameter-efficient-adaptation-of-sam-in-specialized-domains}

Several studies have combined SAM with LoRA or adapter-based fine-tuning outside the natural-image domain. Zhang and Liu \cite{ref10} proposed SAMed for medical image segmentation. Within civil infrastructure, Guo et al.~\cite{ref7} fine-tuned SAM with LoRA for structural crack segmentation, and Ge et al.~\cite{ref8} combined adapter and LoRA fine-tuning in CrackSAM. These are the closest prior studies, but both treat crack segmentation as a single-class problem and both build on the first-generation SAM, whose geometric prompt interface raises none of the class-to-prompt or negative-prompt questions examined here. The present study differs in adapting the concept-prompted third generation, in targeting a multi-class taxonomy, and in ablating adapter placement across two independent datasets.

\subsection{SAM applied to civil infrastructure inspection}\label{sam-applied-to-civil-infrastructure-inspection}

Ahmadi et al.~\cite{ref11} applied SAM zero-shot to civil infrastructure defect assessment, and Ye et al.~\cite{ref12} evaluated SAM-based instance segmentation for structural damage detection on a purpose-built masonry crack dataset. Wu et al.~\cite{ref13} proposed FESS-SAM for full-element semantic segmentation of tunnel linear array images. These studies establish that SAM-family models transfer usefully to infrastructure imagery, but they evaluate zero-shot or fully fine-tuned settings, without a controlled parameter-efficient adaptation study under realistic hardware constraints and without testing whether the adaptation transfers across datasets.

\subsection{Task-specific defect segmentation networks}\label{task-specific-defect-segmentation-networks}

A substantial body of work applies convolutional and encoder--decoder architectures to tunnel lining defect detection without foundation-model pretraining. Dang et al.~\cite{ref14} and Xu et al.~\cite{ref15} proposed pipelines for automatic crack evaluation and mobile-acquisition crack detection. Zhou et al.~\cite{ref16} proposed LC-DeepLab for fast crack detection, Hou et al.~\cite{ref17} an encoder--decoder network for pixel-level crack segmentation, and Li et al.~\cite{ref18} a lightweight Mini-Unet. Wang et al.~\cite{ref19} review classification, detection, and segmentation approaches to shield tunnel lining defects. Water ingress and concrete spalling have received separate treatments: instance segmentation of leakage and scaling defects \cite{ref20,ref21}, structure-from-motion combined with Mask R-CNN \cite{ref22}, mobile laser scanning \cite{ref23}, and SE-TransUNet under complex backgrounds \cite{ref24} for the former; encoder--decoder architectures for spalling detection and severity \cite{ref25,ref26}, machine-vision detection of spalling and exposed rebar \cite{ref27}, and semantic segmentation of concrete defects more broadly \cite{ref28} for the latter. These works demonstrate strong task-specific performance from random or ImageNet initialization, and support the three-category taxonomy adopted here, while treating the categories as separate modeling problems rather than jointly.

\subsection{Benchmarks and cross-dataset generalization}\label{benchmarks-and-cross-dataset-generalization}

Evaluating an adaptation method on a single purpose-built dataset cannot separate a genuinely effective method from one exploiting that dataset's imagery or annotation conventions. S2DS \cite{ref9} provides an independently collected, publicly available concrete-surface benchmark with crack and spalling annotations, and has been used alongside benchmarks such as dacl10k \cite{ref29} for cross-dataset evaluation; Hernandez Noguera et al.~\cite{ref30} used augmented S2DS data for multi-class defect segmentation. We use S2DS as an independent generalization test, applying the identical protocol used on the tunnel dataset.

\section{Method}\label{method}

\subsection{Problem formulation}\label{problem-formulation}

Let a dataset be a collection of images \(I\) with instance annotations \(\{(m_{k}, c_{k})\}\), where \(m_k\) is a binary mask and \(c_k \in \mathcal{C}\) a category drawn from a finite label set. Each category carries a human-readable name, \(\mathrm{name}(c)\).

A conventional segmentation model learns a mapping \(I \mapsto \{(\hat{m}, \hat{c})\}\) in which \(\hat{c}\) ranges over \(\mathcal{C}\), fixed at training time. A concept-promptable model instead realizes

\[f_{\theta}(I, t) \;\longmapsto\; \big(\{(\hat{m}_j, \hat{s}_j)\}_{j=1}^{N},\; \hat{p}\big),\]

where \(t\) is a free-form text prompt, \(\hat{m}_j\) and \(\hat{s}_j\) are the mask and confidence of the \(j\)-th of \(N\) object queries, and \(\hat{p}\) is a scalar presence estimate for whether the prompted concept occurs in \(I\) at all. The label set is not part of the model.

Adaptation therefore has two requirements that a class-labeled dataset does not directly satisfy. It must define a mapping from categories to prompts, \(c \mapsto t\), so that annotations can supervise a language-conditioned model at all (Section~\ref{supervising-a-concept-promptable-model-from-coco-class-labels}). And it must supply supervision for the case \(\hat{p} \to 0\), in which the correct behavior is to return nothing, since a dataset consisting only of annotated objects never exhibits it (Section~\ref{hard-negative-prompt-supervision}). The parameter-efficiency requirement is separate: we seek an update \(\Delta\theta\) with \(|\Delta\theta| \ll |\theta|\) that leaves the pretrained \(\theta\) intact (Section~\ref{low-rank-adaptation-and-adapter-placement}).

\subsection{Architecture}\label{architecture}

\begin{figure*}[!t]
\centering
\includegraphics[width=\textwidth]{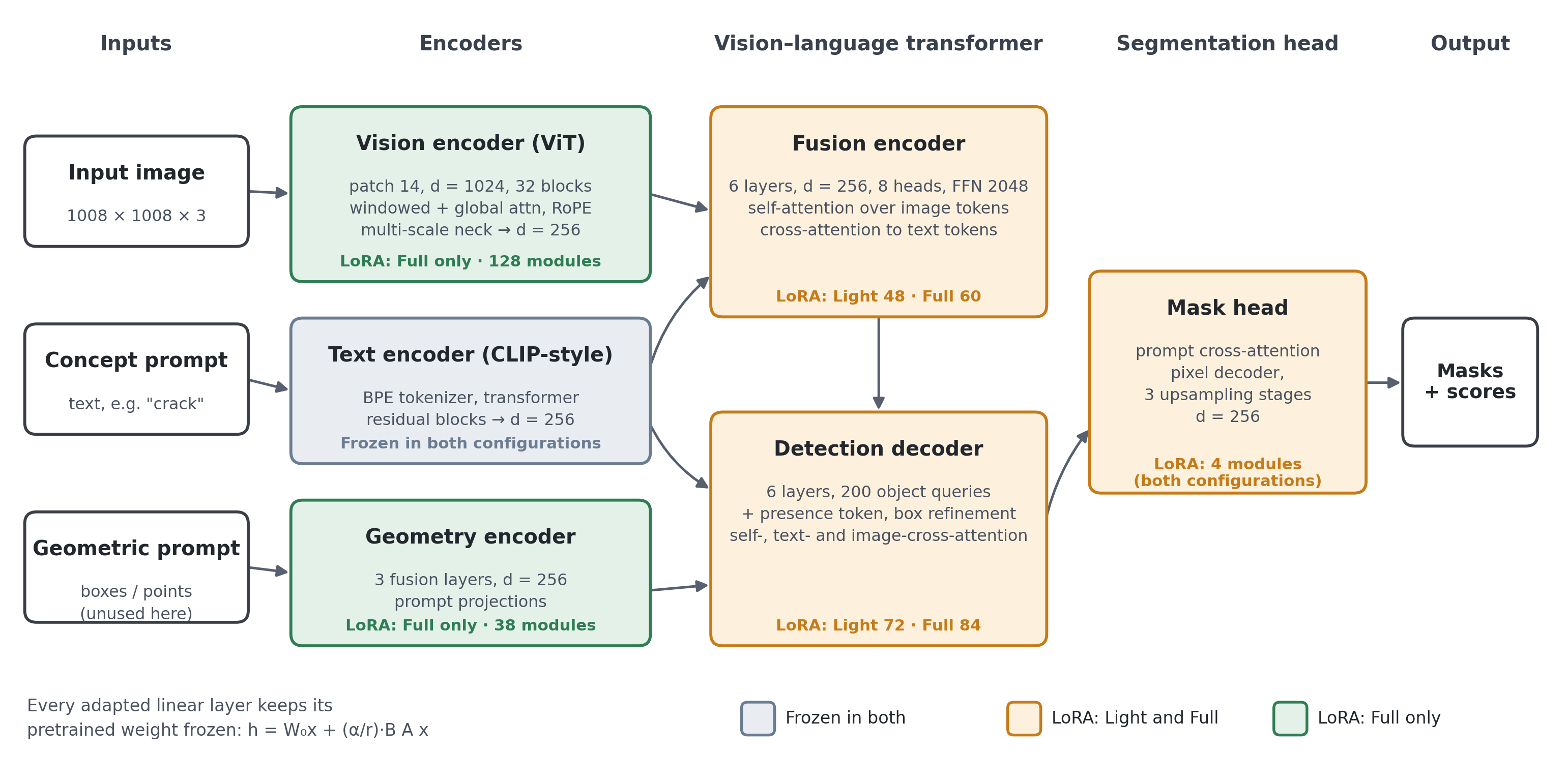}
\caption{SAM3 architecture and adapter placement. Frozen pretrained components are shown in gray, components adapted in both the Light and Full configurations in orange, and components adapted only in the Full configuration in green. Counts give the number of linear layers carrying a LoRA adapter. The text encoder is deliberately frozen in both configurations (Section~\ref{low-rank-adaptation-and-adapter-placement}).}
\label{fig:1}
\end{figure*}

SAM3 comprises five components, shown in Fig.~\ref{fig:1}. The image encoder is a Vision Transformer operating on \(1008 \times 1008\) inputs with patch size 14, embedding width 1024, 32 blocks, and 16 heads; attention is computed within local windows except at four evenly spaced blocks that attend globally, and position is encoded by combined absolute and rotary embeddings. A multi-scale neck projects features to a common width of 256. The text encoder is a CLIP-style transformer with byte-pair-encoding tokenization mapping the prompt to text tokens in the same 256-dimensional space, and a geometry encoder provides an equivalent embedding for box and point prompts, unused in this study. A fusion encoder of six layers interleaves self-attention over image tokens with cross-attention to text tokens, producing prompt-conditioned features. A detection decoder of six layers refines \(N = 200\) learned object queries against those features through self-attention, cross-attention to text, and cross-attention to image tokens, with iterative box refinement; a presence token predicts \(\hat{p}\), decoupling the question of whether the concept is present from the per-query localization scores. A segmentation head cross-attends each query to the prompt embedding and decodes a mask through a pixel decoder with three upsampling stages. The complete model contains approximately 848 million parameters.

\subsection{Low-rank adaptation and adapter placement}\label{low-rank-adaptation-and-adapter-placement}

For a frozen pretrained projection \(\mathbf{W}_0 \in \mathbb{R}^{d_{\text{in}} \times d_{\text{out}}}\), the adapted layer computes

\[\mathbf{h} = \mathbf{W}_0\mathbf{x} + \frac{\alpha}{r}\,\mathbf{B}\mathbf{A}\mathbf{x}, \qquad \mathbf{A} \in \mathbb{R}^{d_{\text{in}} \times r},\;\; \mathbf{B} \in \mathbb{R}^{r \times d_{\text{out}}},\]

where \(r \ll \min(d_{\text{in}}, d_{\text{out}})\) is the adapter rank and \(\alpha\) a scaling factor. Only \(\mathbf{A}\) and \(\mathbf{B}\) are trainable. \(\mathbf{A}\) is initialized from a Kaiming-uniform distribution and \(\mathbf{B}\) to zero, so the adapted model is exactly equivalent to the pretrained model at initialization and departs from it only as the adapters learn; dropout is applied to the adapter input. The trainable parameter count over a set \(\mathcal{L}\) of adapted layers is

\[|\Delta\theta| \;=\; r \sum_{l \in \mathcal{L}} \big(d^{\,l}_{\text{in}} + d^{\,l}_{\text{out}}\big),\]

independent of the magnitude of the frozen weights. Throughout this study \(r = 16\), \(\alpha = 32\), and the adapter dropout probability is 0.1.

One implementation detail is a precondition for adapting attention at all. SAM3's attention blocks use fused multi-head attention, in which the query, key, and value projections are stored as a single concatenated matrix and applied inside a fused kernel, leaving no separate linear layers to adapt. Each such block was re-expressed as an equivalent module holding four explicit projections whose weights are copied from the corresponding slices of the pretrained fused matrix. The re-parameterization is exact, leaves the forward computation and the pretrained values unchanged, and exposes four adaptable projections (query, key, value, and output) per attention block.

Two placements were compared at fixed rank (Table~\ref{tab:1}). The \textbf{Light} configuration adapts the attention projections of the fusion encoder, the detection decoder, and the segmentation head's prompt cross-attention, and nothing else: 124 layers, all of width 256, giving \(124 \times 16 \times (256 + 256) = 1{,}015{,}808\) trainable parameters, or 0.121\% of the model. The \textbf{Full} configuration extends adaptation to the Vision Transformer trunk (the fused query--key--value projection, the attention output projection, and both MLP layers of all 32 blocks), the geometry encoder, and the feed-forward layers of the fusion encoder and decoder, reaching 314 modules and \(11{,}419{,}776\) parameters, or 1.341\%. Because the trunk operates at width 1024 with a feed-forward expansion of 4.625, its 128 adapted layers alone account for approximately 9.0 million of those parameters, which is why a 2.5-fold increase in adapted modules yields an 11.2-fold increase in trainable parameters. Relative to full fine-tuning, both configurations reduce the trainable parameter count by more than 98.6\%.

\begin{table*}[!t]
\centering
\caption{LoRA adapter placement. Counts give the number of linear layers carrying an adapter; a dash indicates a fully frozen component.}
\label{tab:1}
\footnotesize
\setlength{\tabcolsep}{4pt}
\begin{tabularx}{\textwidth}{>{\raggedright\arraybackslash}X>{\raggedright\arraybackslash}Xcc}
\toprule
Component & Adapted linear layers & Light & Full \\
\midrule
Vision encoder (ViT, 32 blocks) & fused query--key--value, attention output projection, MLP layers & --- & 128 \\
Text encoder & none (frozen by design) & --- & --- \\
Geometry encoder (3 layers) & attention projections, feed-forward layers, prompt projections & --- & 38 \\
Fusion encoder (6 layers) & attention projections (self, text cross-attention) & 48 & 48 \\
Fusion encoder (6 layers) & feed-forward layers & --- & 12 \\
Detection decoder (6 layers) & attention projections (self, text cross-, image cross-attention) & 72 & 72 \\
Detection decoder (6 layers) & feed-forward layers & --- & 12 \\
Segmentation head & prompt cross-attention projections & 4 & 4 \\
\textbf{Total adapted modules} &  & \textbf{124} & \textbf{314} \\
\textbf{Trainable parameters} & at rank 16 & \textbf{1,015,808 (0.121\%)} & \textbf{11,419,776 (1.341\%)} \\
\bottomrule
\end{tabularx}
\end{table*}

The text encoder is frozen in both configurations. This is a deliberate design choice rather than a capacity economy. The taxonomy supplies an extremely narrow prompt vocabulary, and adapting the text encoder on it was found to erode the prompt discrimination on which concept prompting depends, namely the model's ability to return nothing when queried with a concept absent from the image. Freezing preserves the pretrained geometry of the concept space while allowing the visual and fusion pathways to specialize on defect appearance; Section~\ref{hard-negative-prompt-supervision} shows that this choice is also what allows negative supervision to generalize beyond the specific negative concepts seen in training.

\subsection{Supervising a concept-promptable model from COCO class labels}\label{supervising-a-concept-promptable-model-from-coco-class-labels}

Both datasets are distributed as COCO-style instance segmentation: one annotation file per split listing the images, a category table, and one record per instance carrying a category identifier, an axis-aligned bounding box in \([x, y, w, h]\) pixel form, and a segmentation field holding either polygon vertex lists or a run-length encoding.

The mapping \(c \mapsto t\) used here is the category table itself. Each category's name string --- ``crack'', ``concrete spalling'', ``water ingress'' --- is lower-cased and used verbatim as the text of a training query. No prompt template (``a photo of a \ldots{}''), no synonym expansion, and no learned class embedding are introduced: the annotation vocabulary \emph{is} the prompt vocabulary. This is viable because applied taxonomies are typically already descriptive noun phrases lying within the pretrained concept space. It is not universally viable: a dataset labeled with opaque codes such as ``C1'' or ``type-2'' would first require a mapping to descriptive names, since a frozen text encoder can contribute nothing to a token it has never meaningfully encountered. The transferability of the procedure is therefore bounded by the linguistic quality of the label set, which is a property of the annotation schema rather than of the model.

Within each image, annotations are grouped by category and one query is emitted per category present, its target set containing every instance of that category. This grouping is what makes the adapted model an exhaustive concept detector rather than a referring-expression segmenter: the query ``crack'' on an image containing four separate cracks is supervised to return all four, not the most salient one. It matters particularly for the tunnel dataset, in which co-located defects of different categories are common (Table~\ref{tab:2}). Categories present in the category table but carrying no annotations anywhere in the split, such as the placeholder supercategory row emitted by some annotation tools, are excluded, since they would define a concept with no visual grounding.

The remaining conversion is geometric. Images are resized to the \(1008 \times 1008\) input resolution and normalized channel-wise; boxes are converted from absolute corner-and-size form to normalized center form; polygon and run-length segmentations are both decoded to binary masks and resampled to the input resolution by nearest-neighbor interpolation. Nearest-neighbor is used in place of bilinear interpolation, which blurs the single-pixel-wide structures that dominate crack annotations. An image carrying no annotations is not discarded but contributes a single generic query with an empty target set, so that unannotated background still supervises the model.

\subsection{Hard-negative prompt supervision}\label{hard-negative-prompt-supervision}

A dataset converted as above yields positive prompts exclusively: every query issued during training has at least one matching object. Fine-tuning an open-vocabulary detector under that regime produces a specific and easily overlooked collapse. Because the model is never shown a prompt that should return nothing, the presence estimate \(\hat{p}\) decouples from the text condition and degenerates into always firing, so that after adaptation on crack, concrete spalling, and water ingress, prompting with an unrelated concept such as ``car'' still returns the crack. Detection accuracy on the trained categories improves while the discrimination that makes a promptable model useful is destroyed. The collapse is invisible to every metric computed on positive prompts, including all ten reported in Section~\ref{evaluation-metrics}, which is what makes it worth stating explicitly: a study that did not test for it would not observe it.

Hard-negative text queries address this directly. A negative query is issued through the same mechanism as a positive one but with an empty target set, so that all \(N = 200\) object queries are supervised toward the no-object state and the presence token toward absence. Two tiers were used, and the distinction between them matters.

The first tier is \textbf{in-domain and exhaustive}: for every image, each dataset category not annotated in that image is issued as a zero-detection query. On a water-ingress image, ``crack'' and ``concrete spalling'' are each supervised to return nothing, on every pass through the data. These are the confusable prompts, and the supervision costs nothing in annotation effort, being derived entirely from the existing label set. Sampling these negatives randomly rather than including them exhaustively was found to provide too sparse a signal to separate visually similar categories: under random sampling, all three defect prompts converged on segmenting the same region. For the three-category tunnel taxonomy this tier contributes at most two negative queries per annotated image; on S2DS, where every image is single-category, it contributes exactly one.

The second tier is \textbf{out-of-domain and sampled}: up to three concepts per image are drawn from a pool of twenty generic everyday objects such as ``car'', ``person'', and ``bicycle''. One safeguard governs the pool. Any entry sharing a word with a dataset category is removed automatically, so ``water'' is dropped once a category named ``water ingress'' exists, since supervising ``water returns nothing'' on images of water ingress would contradict the positive class. The same consideration constrains any reuse of the procedure: a pool concept that can genuinely appear in the imagery, such as ``road'' for pavement defect data, must be replaced rather than trained against.

Because the text encoder is frozen (Section~\ref{low-rank-adaptation-and-adapter-placement}), what these negatives retrain is a general decision rule --- fire only when the prompt embedding matches the visual evidence --- rather than a blocklist of specific strings. The pretrained encoder already places ``plane'' near ``car'' and far from ``crack'', and freezing preserves that geometry, so a prompt never seen as a training negative is expected to return nothing as well. The known boundary of this generalization is near-synonyms of the target categories: prompts such as ``fracture'' or ``gap'' lie close to ``crack'' in the frozen concept space and may continue to fire, which is the model reporting a genuine conceptual proximity rather than a defect in the supervision. Where a specific neighboring term must be suppressed, it must be added to the pool explicitly.

Negative queries are constructed for the validation split as well as for training, so the validation loss driving early stopping reflects prompt discrimination and not only mask quality on positive prompts.

\subsection{Training objective}\label{training-objective}

The objective is SAM3's detection-and-segmentation loss, applied identically to positive and negative queries. Predictions from the 200 object queries are assigned to ground-truth instances by Hungarian bipartite matching under a cost combining classification (weight 2.0), bounding-box \(L_1\) distance (5.0), and generalized intersection-over-union (2.0). Matched predictions are penalized by a binary classification loss with positive-class weight 10 together with a presence loss (each weighted 20), box \(L_1\) and generalized-IoU losses (5 and 2), and mask losses comprising a focal term with \(\alpha = 0.25\), \(\gamma = 2\) (weight 200) and a Dice term (weight 10). The high relative weight on the mask terms reflects that pixel-level mask quality, rather than box localization, is the quantity of interest for defect segmentation. An auxiliary one-to-many matcher, permitting a ground-truth instance to be matched by up to four queries, contributes a further set of terms at twice the base weight and densifies the training signal for the elongated, frequently fragmented shapes typical of cracks.

\section{Experimental Setup}\label{experimental-setup}

\subsection{Datasets}\label{datasets}

Two datasets were used. The tunnel lining dataset is purpose-built and comprises three defect categories, with co-located annotations of different categories common within a single image. Its training split (431 original images) was augmented three-fold (horizontal flip, brightness, exposure) to 1,292 images yielding 2,407 training annotations, of which crack accounts for 70.8\%. S2DS \cite{ref9} contains two overlapping categories with no water-ingress equivalent, is single-category per image throughout, and was used at native scale without augmentation. The two datasets share no images or annotations. Table~\ref{tab:2} summarizes the comparison; the structural differences between them, particularly multi-defect co-occurrence and augmentation, are relevant to interpreting the cross-dataset results in Section~\ref{discussion}.

\begin{table*}[!t]
\centering
\caption{Structural comparison of the two evaluation datasets.}
\label{tab:2}
\footnotesize
\setlength{\tabcolsep}{4pt}
\begin{tabularx}{\textwidth}{>{\raggedright\arraybackslash}X>{\raggedright\arraybackslash}X>{\raggedright\arraybackslash}X}
\toprule
Characteristic & Tunnel lining & S2DS \\
\midrule
Defect categories & 3 (crack, concrete spalling, water ingress) & 2 (crack, spalling); no water-ingress equivalent \\
Multi-defect images & Common & None (every image is single-category) \\
Train images (original / augmented) & 431 / 1,292 & 563 / 563 (no augmentation) \\
Validation images & 81 & 87 \\
Test images & 50 & 93 \\
Train annotations & 2,407 & 1,243 \\
Majority training category & Crack (70.8\%) & Crack (70.9\%) \\
Image domain & Underground tunnel segment lining & General concrete-surface benchmark \\
\bottomrule
\end{tabularx}
\end{table*}

\subsection{Training configuration}\label{training-configuration}

All models were trained on a workstation with two NVIDIA RTX 5090 GPUs (32,607 MiB each), CUDA 13.0, and PyTorch 2.13.0. Adapter rank, dropout, and weight decay were held constant across configurations at 16, 0.1, and 0.01, with a learning rate of \(1 \times 10^{-4}\) for both Light and Full. On the tunnel dataset both configurations were trained at batch size 2 across two GPUs; on S2DS, Full required distributed training at batch size 1 per GPU (effective batch size 2) owing to the memory cost of its larger adapted footprint. Optimization used AdamW over the adapter matrices only, a cosine learning-rate schedule with 200 warmup steps, gradient clipping at norm 1.0, gradient accumulation over eight steps, bfloat16 mixed precision, and a fixed random seed. Early stopping on validation loss was applied throughout. Checkpoints store adapter weights alone, which keeps each saved model in the tens of megabytes rather than the several gigabytes of a full checkpoint --- a practical property for organizations maintaining several site-specific adaptations of one base model.

The frozen baseline is SAM3 with no adaptation, prompted with the same category names and evaluated through the identical pipeline, isolating the effect of adaptation from every other factor.

\subsection{Evaluation metrics}\label{evaluation-metrics}

Ten metrics spanning detection and segmentation quality were computed independently on the validation and test splits of each dataset: mean average precision across IoU thresholds 0.50--0.95 (mAP), and at 0.50 and 0.75 (mAP@50, mAP@75); category-grouped F1 over the same range (cgF1) and at 0.50 (cgF1@50); pixel-level IoU, precision, and recall; mean per-prompt IoU; and instance-level (IL) recall. Pixel-level metrics are computed per (image, prompt) unit by taking the union of all predicted masks and comparing it pixel-wise against the union of ground-truth masks at \(288 \times 288\) mask resolution; the counts denoted \(n\) (units) in the per-category tables are numbers of such units. Predictions are gated by the presence estimate before scoring.

Reporting both families is necessary rather than redundant, for a reason specific to thin structures. Instance matching requires one prediction to cover one ground-truth instance at IoU above threshold. When a model predicts a single crack as two disjoint portions --- a common and visually benign outcome --- each portion's IoU against the full ground truth is at best about 0.5, since the intersection is only its own pixels while the union is the whole crack. The ground truth is therefore counted as missed \emph{and} both portions are counted as false positives, even though together they cover the crack accurately. This triple penalty is a principal reason mAP@75 falls far below mAP@50 for elongated structures, and it means that instance-level metrics systematically understate segmentation quality in exactly the category that dominates both datasets. Pixel-level metrics, which are insensitive to fragmentation and standard in the crack-segmentation literature, are reported alongside them for this reason; a large gap between pixel IoU and mAP@50 indicates a model that finds the cracks but splits them.

\subsection{Evaluation integrity}\label{evaluation-integrity}

During initial S2DS evaluation, an indexing error in the construction of ground truth within the validation script assigned each prediction's source-image reference to the raw one-indexed COCO image identifier rather than the corresponding zero-indexed dataset position, so that every prediction was scored against an incorrect ground-truth mask. This produced near-zero detection and pixel metrics despite the same checkpoints generating visibly accurate predictions under a separate single-image inference path. Correcting the error changed pixel IoU from approximately 0.02 to approximately 0.86 with no change to model weights or predictions, confirming a scoring artifact rather than a modeling deficiency. A related hardcoded single-category placeholder in the same routine was corrected in the same pass, enabling the per-category breakdown of Section~\ref{per-category-results}. All S2DS results reported here reflect runs performed after both corrections. We report this explicitly because the failure signature --- uniformly near-zero metrics against qualitatively correct predictions --- is a diagnostic that recurs in evaluation pipelines for promptable models, where predictions and ground truth are associated through prompt-conditioned rather than image-indexed structures.

\section{Results}\label{results}

\subsection{Aggregate performance}\label{aggregate-performance}

On the tunnel test split, Full LoRA achieved the highest pixel precision (0.519) and instance-level recall (0.672), while Light LoRA achieved the highest pixel IoU (0.338) and pixel recall (0.589). Both configurations substantially outperformed the frozen baseline, which reached a test pixel IoU of 0.017 and instance-level recall of 0.375. On S2DS, Light LoRA achieved the highest test pixel IoU (0.8554) and mean per-prompt IoU (0.5831), and both configurations reached perfect instance-level recall (1.000) against 0.574 for the baseline. Tables~\ref{tab:3} and~\ref{tab:4} report the full suite.

\begin{table*}[!t]
\centering
\caption{Tunnel lining detection and segmentation metrics.}
\label{tab:3}
\footnotesize
\setlength{\tabcolsep}{4pt}
\begin{tabularx}{\textwidth}{lcccccc}
\toprule
Metric & Baseline (valid) & Baseline (test) & Light (valid) & Light (test) & Full (valid) & Full (test) \\
\midrule
mAP (0.50:0.95) & 0.023 & 0.006 & 0.068 & 0.048 & \textbf{0.085} & \textbf{0.053} \\
mAP@50 & 0.102 & 0.043 & 0.178 & 0.107 & \textbf{0.205} & \textbf{0.109} \\
mAP@75 & 0.001 & 0.0002 & 0.052 & \textbf{0.046} & \textbf{0.064} & 0.045 \\
cgF1 (0.50:0.95) & 0.026 & 0.004 & 0.102 & 0.034 & \textbf{0.110} & \textbf{0.037} \\
cgF1@50 & 0.092 & 0.023 & 0.222 & 0.069 & \textbf{0.233} & \textbf{0.074} \\
Pixel IoU & 0.092 & 0.017 & 0.579 & \textbf{0.338} & \textbf{0.586} & 0.329 \\
Pixel precision & 0.136 & 0.018 & \textbf{0.732} & 0.441 & 0.674 & \textbf{0.519} \\
Pixel recall & 0.219 & 0.303 & 0.735 & \textbf{0.589} & \textbf{0.818} & 0.472 \\
Mean per-prompt IoU & 0.161 & 0.096 & 0.391 & 0.270 & \textbf{0.413} & \textbf{0.282} \\
IL recall & 0.429 & 0.375 & \textbf{0.804} & 0.641 & 0.786 & \textbf{0.672} \\
\bottomrule
\end{tabularx}
\vspace{2pt}
\parbox{\textwidth}{\footnotesize \textit{Note.} Bold marks the better of the two adapted configurations, Light and Full, within each split. The frozen baseline is reported for reference and is not ranked. Higher is better for every metric. mAP is mean average precision, cgF1 category-grouped F1, and IL recall instance-level recall.}
\end{table*}

\begin{table*}[!t]
\centering
\caption{S2DS detection and segmentation metrics.}
\label{tab:4}
\footnotesize
\setlength{\tabcolsep}{4pt}
\begin{tabularx}{\textwidth}{lcccccc}
\toprule
Metric & Baseline (valid) & Baseline (test) & Light (valid) & Light (test) & Full (valid) & Full (test) \\
\midrule
mAP (0.50:0.95) & 0.0000 & 0.0000 & 0.1121 & 0.0942 & \textbf{0.1272} & \textbf{0.0978} \\
mAP@50 & 0.0000 & 0.0000 & 0.1897 & \textbf{0.2207} & \textbf{0.1985} & 0.2124 \\
mAP@75 & 0.0000 & 0.0000 & 0.1101 & \textbf{0.0831} & \textbf{0.1256} & 0.0825 \\
cgF1 (0.50:0.95) & 0.0000 & 0.0000 & \textbf{0.1793} & \textbf{0.1349} & 0.1117 & 0.1115 \\
cgF1@50 & 0.0000 & 0.0000 & \textbf{0.2805} & \textbf{0.2822} & 0.1751 & 0.2317 \\
Pixel IoU & 0.0150 & 0.0165 & \textbf{0.8665} & \textbf{0.8554} & 0.8526 & 0.8363 \\
Pixel precision & 0.0973 & 0.1217 & \textbf{0.9331} & \textbf{0.9185} & 0.9161 & 0.9003 \\
Pixel recall & 0.0174 & 0.0187 & 0.9239 & \textbf{0.9257} & \textbf{0.9249} & 0.9217 \\
Mean per-prompt IoU & 0.0808 & 0.0931 & \textbf{0.5433} & \textbf{0.5831} & 0.5361 & 0.5711 \\
IL recall & 0.521 & 0.574 & \textbf{0.937} & \textbf{1.000} & 0.896 & \textbf{1.000} \\
\bottomrule
\end{tabularx}
\vspace{2pt}
\parbox{\textwidth}{\footnotesize \textit{Note.} Bold marks the better of the two adapted configurations, Light and Full, within each split. The frozen baseline is reported for reference and is not ranked. Higher is better for every metric. Metric abbreviations are as in Table~\ref{tab:3}.}
\end{table*}

\subsection{Cross-dataset consistency of the improvement}\label{cross-dataset-consistency-of-the-improvement}

Table~\ref{tab:5} reports the improvement of the best configuration over the frozen baseline on the test split of each dataset. Every metric with a non-degenerate baseline improved by between 1.7-fold and 51.8-fold on at least one dataset, and improvements were directionally consistent for every metric computable on both. Pixel IoU improved 19.9-fold on the tunnel dataset and 51.8-fold on S2DS; pixel precision 28.8-fold and 7.5-fold; pixel recall 1.9-fold and 49.5-fold; mean per-prompt IoU 2.9-fold and 6.3-fold; and instance-level recall 1.8-fold and 1.7-fold, the closest-matching pair.

For the five S2DS detection metrics the frozen baseline scored exactly zero, detecting no true positives at any confidence or IoU threshold, so no finite multiplier is defined. Both adapted configurations nonetheless reached absolute detection values (test mAP 0.094--0.098, cgF1 0.112--0.135) of the same order as, and in fact above, the corresponding tunnel results (test mAP 0.048--0.053, cgF1 0.034--0.037).

\begin{table*}[!t]
\centering
\caption{Improvement of the best configuration over the frozen baseline (test split). A dash marks cells where the baseline is exactly zero and no finite multiplier is defined.}
\label{tab:5}
\footnotesize
\setlength{\tabcolsep}{4pt}
\begin{tabularx}{\textwidth}{lcccccc}
\toprule
Metric (test) & Tunnel base & Tunnel best & Fold & S2DS base & S2DS best & Fold \\
\midrule
mAP (0.50:0.95) & 0.006 & 0.053 (Full) & 8.8$\times$ & 0.0000 & 0.0978 (Full) & --- \\
mAP@50 & 0.043 & 0.109 (Full) & 2.5$\times$ & 0.0000 & 0.2207 (Light) & --- \\
mAP@75 & 0.0002 & 0.046 (Light) & 230$\times$* & 0.0000 & 0.0831 (Light) & --- \\
cgF1 (0.50:0.95) & 0.004 & 0.037 (Full) & 9.3$\times$ & 0.0000 & 0.1349 (Light) & --- \\
cgF1@50 & 0.023 & 0.074 (Full) & 3.2$\times$ & 0.0000 & 0.2822 (Light) & --- \\
Pixel IoU & 0.017 & 0.338 (Light) & 19.9$\times$ & 0.0165 & 0.8554 (Light) & 51.8$\times$ \\
Pixel precision & 0.018 & 0.519 (Full) & 28.8$\times$ & 0.1217 & 0.9185 (Light) & 7.5$\times$ \\
Pixel recall & 0.303 & 0.589 (Light) & 1.9$\times$ & 0.0187 & 0.9257 (Light) & 49.5$\times$ \\
Mean per-prompt IoU & 0.096 & 0.282 (Full) & 2.9$\times$ & 0.0931 & 0.5831 (Light) & 6.3$\times$ \\
IL recall & 0.375 & 0.672 (Full) & 1.8$\times$ & 0.574 & 1.000 (both) & 1.7$\times$ \\
\bottomrule
\end{tabularx}
\vspace{2pt}
\parbox{\textwidth}{\footnotesize \textit{Note.} Fold is the ratio of the best configuration to the frozen baseline, with the winning configuration named in parentheses. The starred value rests on a near-zero baseline denominator (0.0002) and should be read as directionally consistent rather than as a precise multiplier.}
\end{table*}

\subsection{Per-category results}\label{per-category-results}

Tables~\ref{tab:6} and~\ref{tab:7} disaggregate pixel metrics by category. The dominant pattern is that the benefit of adaptation is inversely related to the frozen model's prior competence on a category, and that this holds on both datasets.

On the tunnel dataset, the frozen baseline achieved a test pixel IoU of 0.199 on crack, 0.011 on concrete spalling, and exactly 0.000 on water ingress. After adaptation the ordering is compressed and partly reversed: 0.337 for crack (Full), 0.395 for spalling (Light), and 0.378 for water ingress (Light), corresponding to gains of 1.7-fold and 35.9-fold, and from zero for water ingress. On S2DS the same relationship holds with the categories exchanged: the baseline scored 0.1533 on crack and exactly 0.0000 on spalling, while after adaptation spalling reached 0.8720 (Light) and crack 0.4317 (Full), an unbounded gain and a 2.8-fold gain respectively.

Crack is the majority training category in both datasets (70.8\% and 70.9\% of annotations) and yet attains the lowest post-adaptation pixel IoU in both. Its behavior is also distinctive in the frozen model: baseline pixel recall on crack is high (0.720 tunnel, 0.9911 S2DS) while baseline IoU is low, indicating broad, imprecise predictions that cover the crack along with a large surrounding area. Section~\ref{thin-structure-geometry-and-the-choice-of-metric} relates this to the geometry of thin structures.

Both datasets agree that the frozen baseline scored exactly zero pixel IoU and zero recall on the category with no natural-image analogue in the pretraining distribution: water ingress on the tunnel dataset, spalling on S2DS.

\begin{table*}[!t]
\centering
\caption{Tunnel per-category pixel metrics. $n$ is the number of (image, prompt) evaluation units.}
\label{tab:6}
\footnotesize
\setlength{\tabcolsep}{4pt}
\begin{tabularx}{\textwidth}{llcccccc}
\toprule
Category & Metric & Baseline (valid) & Baseline (test) & Light (valid) & Light (test) & Full (valid) & Full (test) \\
\midrule
Crack & Pixel IoU & 0.416 & 0.199 & 0.562 & 0.226 & \textbf{0.629} & \textbf{0.337} \\
Crack & Pixel recall & 0.614 & 0.720 & 0.798 & \textbf{0.794} & \textbf{0.819} & 0.782 \\
Crack & $n$ (units) & 28 & 31 & 28 & 31 & 28 & 31 \\
Concrete spalling & Pixel IoU & 0.032 & 0.011 & \textbf{0.410} & \textbf{0.395} & 0.376 & 0.314 \\
Concrete spalling & Pixel recall & 0.309 & 0.295 & \textbf{0.609} & \textbf{0.588} & 0.472 & 0.401 \\
Concrete spalling & $n$ (units) & 19 & 28 & 19 & 28 & 19 & 28 \\
Water ingress & Pixel IoU & 0.000 & 0.000 & \textbf{0.646} & \textbf{0.378} & 0.612 & 0.363 \\
Water ingress & Pixel recall & 0.000 & 0.000 & 0.736 & 0.433 & \textbf{0.906} & \textbf{0.443} \\
Water ingress & $n$ (units) & 9 & 5 & 9 & 5 & 9 & 5 \\
\bottomrule
\end{tabularx}
\vspace{2pt}
\parbox{\textwidth}{\footnotesize \textit{Note.} Bold marks the better of the two adapted configurations, Light and Full, within each split. The frozen baseline is reported for reference and is not ranked. Higher is better for every metric. The $n$ rows count (image, prompt) evaluation units and are not ranked.}
\end{table*}

\begin{table*}[!t]
\centering
\caption{S2DS per-category pixel metrics.}
\label{tab:7}
\footnotesize
\setlength{\tabcolsep}{4pt}
\begin{tabularx}{\textwidth}{llcccccc}
\toprule
Category & Metric & Baseline (valid) & Baseline (test) & Light (valid) & Light (test) & Full (valid) & Full (test) \\
\midrule
Spalling & Pixel IoU & 0.0000 & 0.0000 & \textbf{0.8848} & \textbf{0.8720} & 0.8702 & 0.8510 \\
Spalling & Pixel recall & 0.0000 & 0.0000 & 0.9258 & \textbf{0.9263} & \textbf{0.9269} & 0.9225 \\
Spalling & $n$ (units) & 23 & 20 & 23 & 20 & 23 & 20 \\
Crack & Pixel IoU & 0.1093 & 0.1533 & \textbf{0.3758} & 0.4242 & 0.3736 & \textbf{0.4317} \\
Crack & Pixel recall & 0.9840 & 0.9911 & \textbf{0.8175} & \textbf{0.8966} & 0.8113 & 0.8787 \\
Crack & $n$ (units) & 25 & 27 & 25 & 27 & 25 & 27 \\
\bottomrule
\end{tabularx}
\vspace{2pt}
\parbox{\textwidth}{\footnotesize \textit{Note.} Bold marks the better of the two adapted configurations, Light and Full, within each split. The frozen baseline is reported for reference and is not ranked. Higher is better for every metric. The $n$ rows count (image, prompt) evaluation units and are not ranked.}
\end{table*}

\subsection{Adapter capacity}\label{adapter-capacity}

Table~\ref{tab:8} summarizes the configurations compared on each dataset. On the tunnel dataset, Full favored detection-oriented metrics (mAP, pixel precision, instance-level recall) while Light favored pixel-level generalization (pixel IoU, pixel recall). On S2DS the two were much closer, with Light holding a small but consistent pixel-IoU advantage and no clear detection advantage accruing to Full.

An important limitation attaches to the cross-dataset half of this comparison. The tunnel and S2DS experiments were run on different revisions of the implementation, and the resulting adapter scopes are not equivalent despite the shared configuration labels: the tunnel Light configuration adapts 124 attention projections across the fusion encoder, decoder, and segmentation head, whereas the S2DS Light configuration adapts 67 modules comprising vision-backbone MLP layers plus one mask-decoder attention module. The Light-versus-Full comparison is therefore internally valid within each dataset but cannot be pooled across them; Section~\ref{adapter-capacity-and-a-confound} treats this as a confound.

\begin{table*}[!t]
\centering
\caption{Adapter configurations compared on each dataset.}
\label{tab:8}
\footnotesize
\setlength{\tabcolsep}{4pt}
\begin{tabularx}{\textwidth}{llc>{\raggedright\arraybackslash}Xl}
\toprule
Dataset & Configuration & Rank & Adapted modules & Trainable parameters \\
\midrule
Tunnel & Light & 16 & 124 (fusion encoder, decoder, mask-head attention) & 1,015,808 (0.121\%) \\
Tunnel & Full & 16 & 314 (adds ViT trunk, geometry encoder, FFN layers) & 11,419,776 (1.341\%) \\
S2DS & Light & 16 & 67 (vision-backbone MLP + 1 mask-decoder attention) & 5,922,816 (0.700\%) \\
S2DS & Full & 16 & 188 (vision-backbone MLP + fusion encoder/decoder) & 6,914,048 (0.816\%) \\
\bottomrule
\end{tabularx}
\end{table*}

\subsection{Training dynamics}\label{training-dynamics}

\begin{figure*}[!t]
\centering
\includegraphics[width=\textwidth]{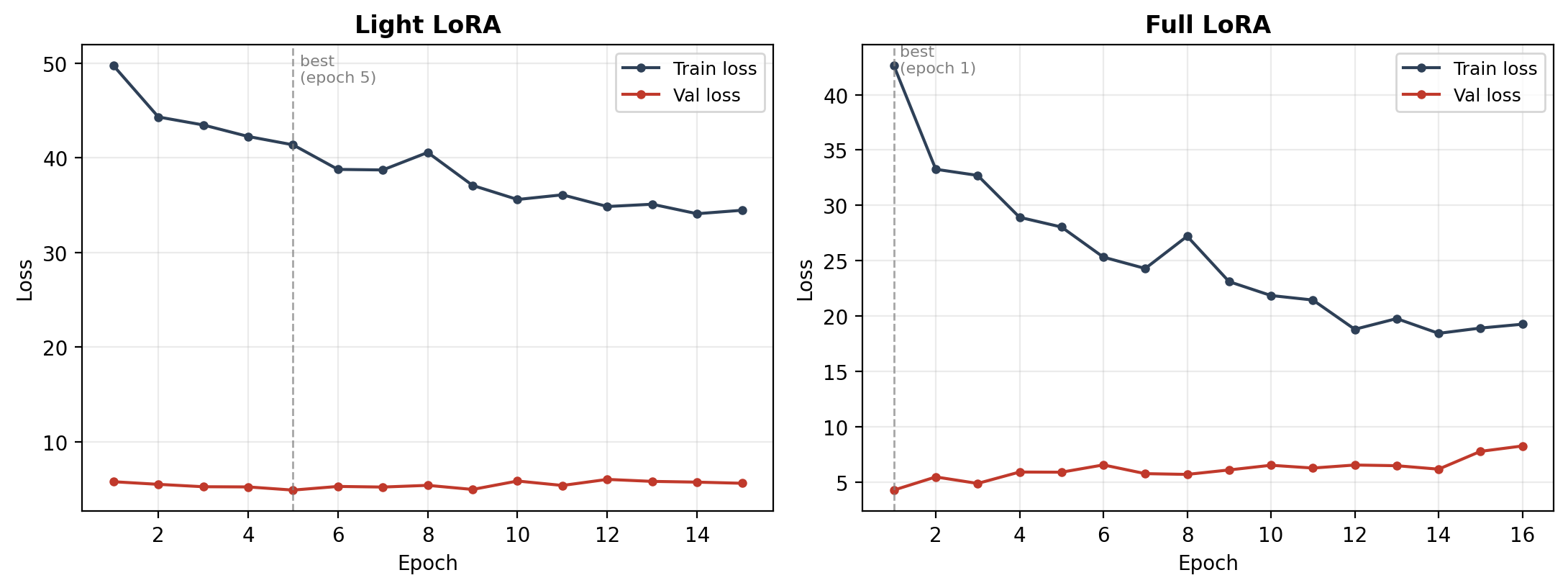}
\caption{Tunnel lining training and validation loss for Light (left, best checkpoint epoch 5) and Full (right, best checkpoint epoch 1). Dashed lines mark each configuration's best checkpoint.}
\label{fig:2}
\end{figure*}

On both datasets the higher-capacity configuration reached its validation-loss minimum earlier and diverged more rapidly thereafter (Table~\ref{tab:9}, Fig.~\ref{fig:2}), consistent with a larger adapted footprint fitting the training distribution faster at the cost of a shorter useful training window. On the tunnel dataset, Full reached its minimum at epoch 1 and rose from 4.30 to 8.28 by epoch 16, whereas Light reached its minimum at epoch 5 and degraded gradually. The pattern recurs on S2DS with epochs 3 and 7 respectively. Given training sets of a few hundred to a few thousand images, this argues for frequent validation and early stopping as a default rather than an optimization, particularly for larger adapter footprints.

\begin{table*}[!t]
\centering
\caption{Best-checkpoint epoch and overfitting behavior.}
\label{tab:9}
\footnotesize
\setlength{\tabcolsep}{4pt}
\begin{tabularx}{\textwidth}{llcc>{\raggedright\arraybackslash}X}
\toprule
Dataset & Configuration & Best epoch & Best validation loss & Overfitting onset \\
\midrule
Tunnel & Light & 5 & $\approx$ 4.87 & Gradual (epochs 5--15: 4.87 $\rightarrow$ 5.59) \\
Tunnel & Full & 1 & $\approx$ 4.30 & Immediate (epochs 1--16: 4.30 $\rightarrow$ 8.28) \\
S2DS & Light & 7 & 4.267 & Gradual and mild (epochs 7--15: 4.27 $\rightarrow$ 4.27) \\
S2DS & Full & 3 & 3.597 & Early (epochs 3--11: 3.60 $\rightarrow$ 3.78, early-stopped) \\
\bottomrule
\end{tabularx}
\end{table*}

\subsection{Qualitative results}\label{qualitative-results}

\begin{figure*}[!t]
\centering
\includegraphics[width=\textwidth]{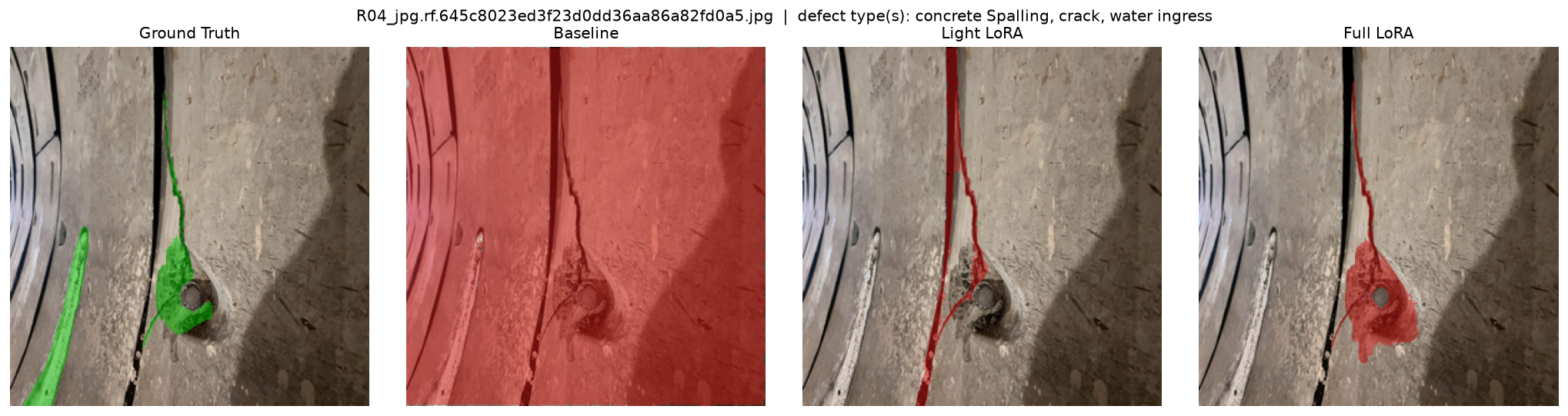}
\caption{Tunnel test image R04 (concrete spalling, crack, water ingress; multi-defect). The baseline floods nearly the entire frame with false-positive defect area, while Light and Full both localize the two true defect regions.}
\label{fig:3}
\end{figure*}

\begin{figure*}[!t]
\centering
\includegraphics[width=\textwidth]{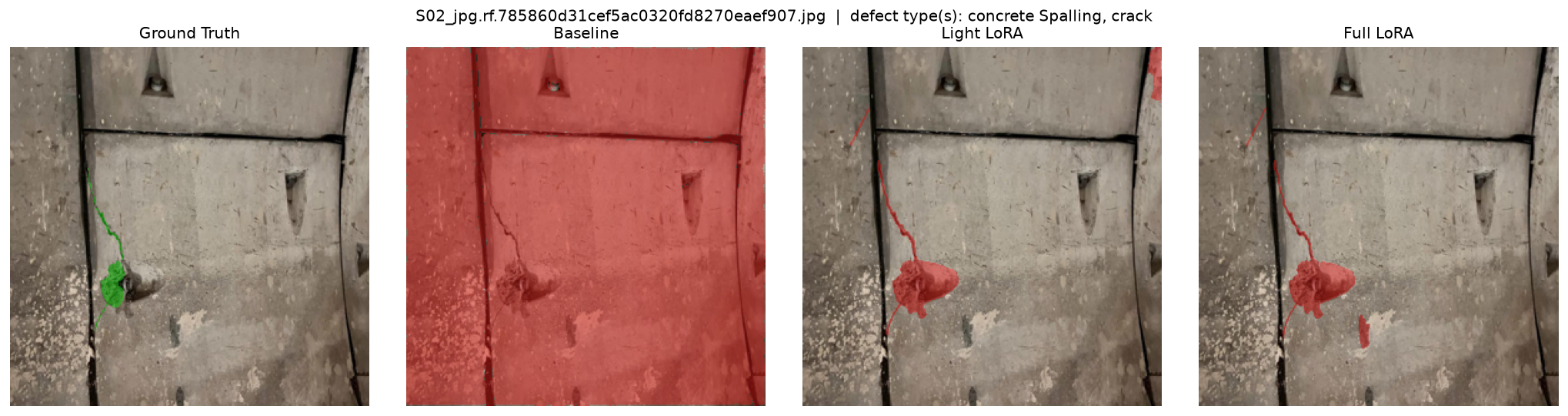}
\caption{Tunnel test image S02 (concrete spalling, crack). A second instance of the baseline's large-area over-prediction, resolved by both adapted configurations.}
\label{fig:4}
\end{figure*}

\begin{figure*}[!t]
\centering
\includegraphics[width=\textwidth]{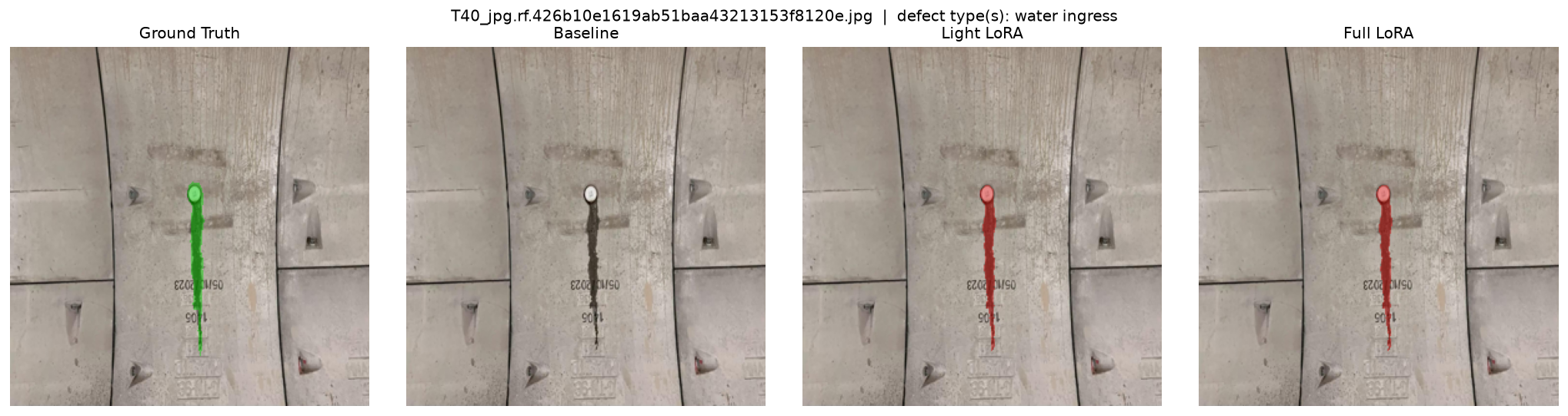}
\caption{Tunnel test image T40 (water ingress). Light and Full both segment the water-ingress streak closely, while the baseline shows only a faint, low-confidence trace.}
\label{fig:5}
\end{figure*}

\begin{figure*}[!t]
\centering
\includegraphics[width=\textwidth]{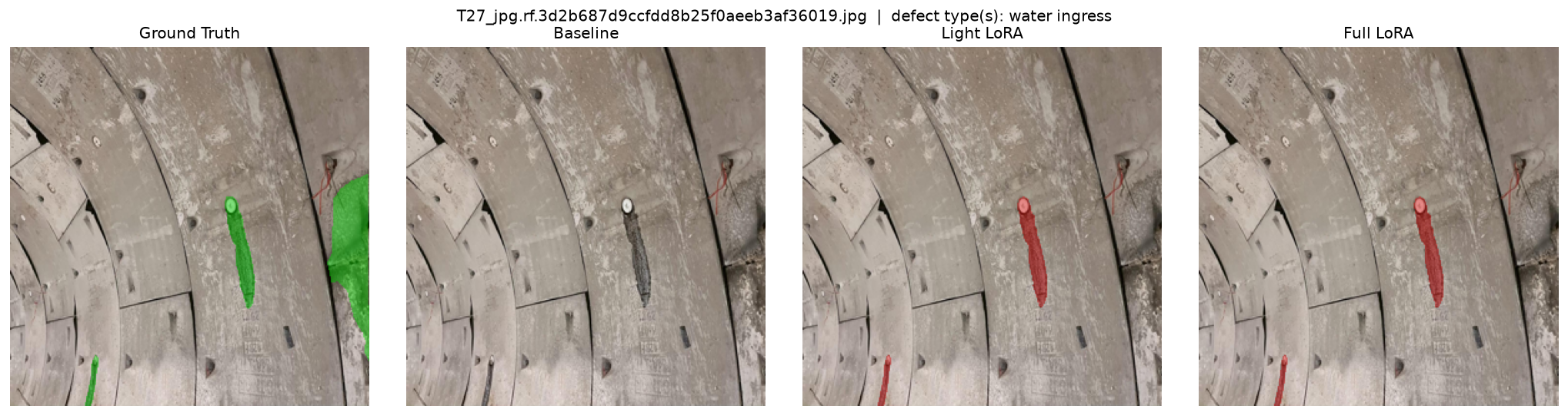}
\caption{Tunnel test image T27 (water ingress). A second water-ingress example showing the same pattern as Fig.~\ref{fig:5}.}
\label{fig:6}
\end{figure*}

\begin{figure*}[!t]
\centering
\includegraphics[width=\textwidth]{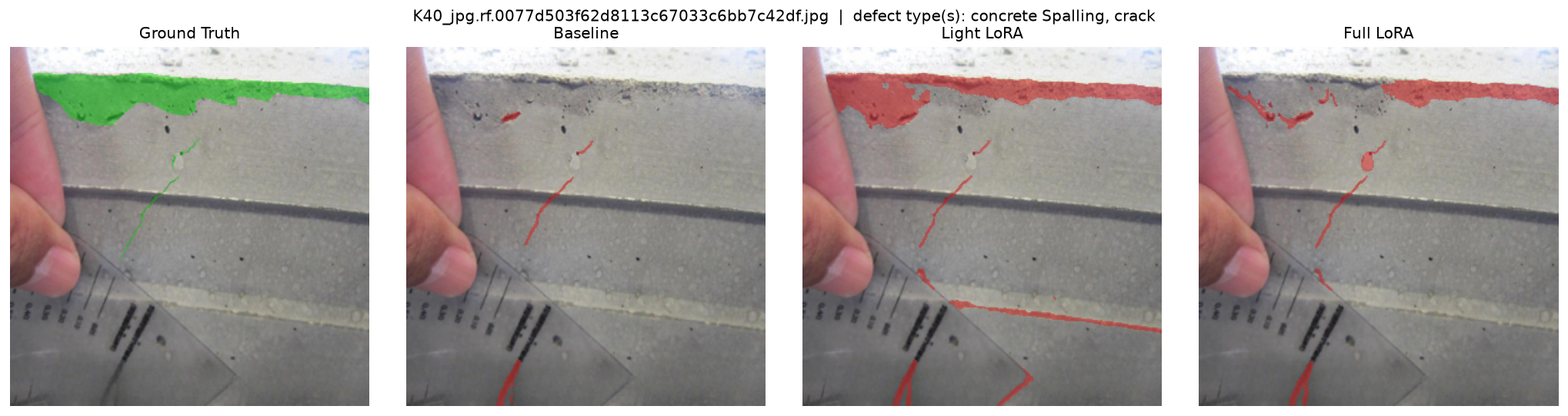}
\caption{Tunnel test image K40 (concrete spalling, crack). Light produces a contiguous, well-bounded spalling prediction; Full's prediction is more fragmented.}
\label{fig:7}
\end{figure*}

\begin{figure*}[!t]
\centering
\includegraphics[width=\textwidth]{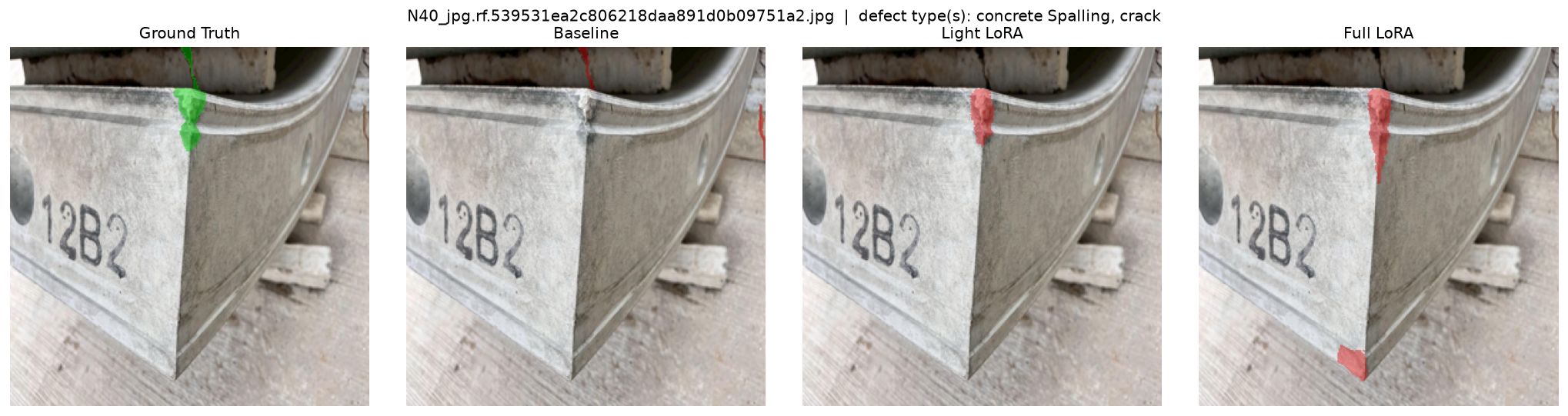}
\caption{Tunnel test image N40 (concrete spalling, crack). Light matches the ground truth closely; Full adds a spurious prediction in an unrelated region.}
\label{fig:8}
\end{figure*}

\begin{figure*}[!t]
\centering
\includegraphics[width=\textwidth]{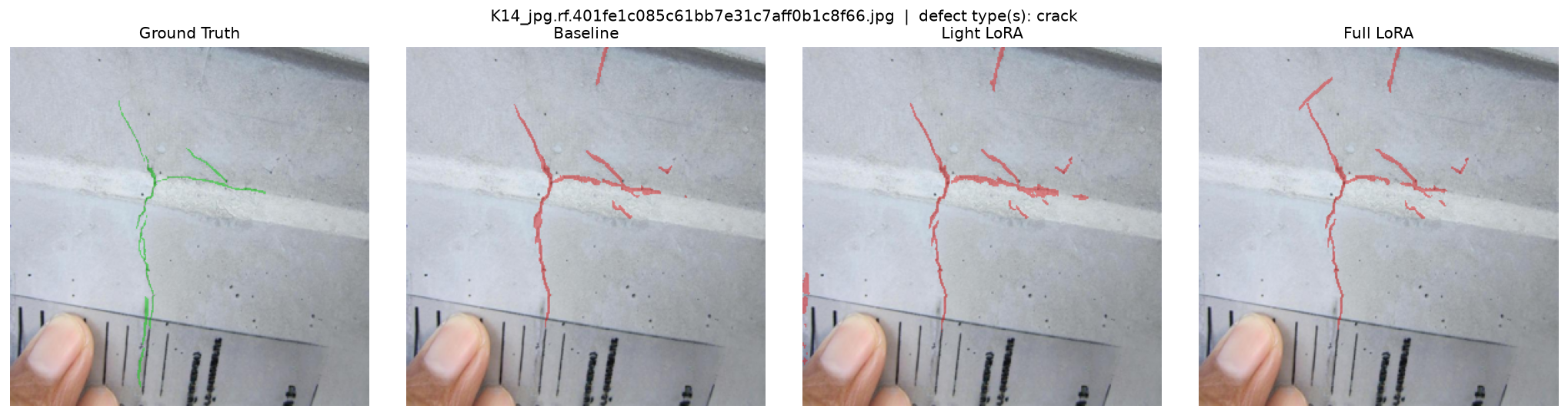}
\caption{Tunnel test image K14 (crack). All three models, including the frozen baseline, track the branching crack pattern closely.}
\label{fig:9}
\end{figure*}

\begin{figure*}[!t]
\centering
\includegraphics[width=\textwidth]{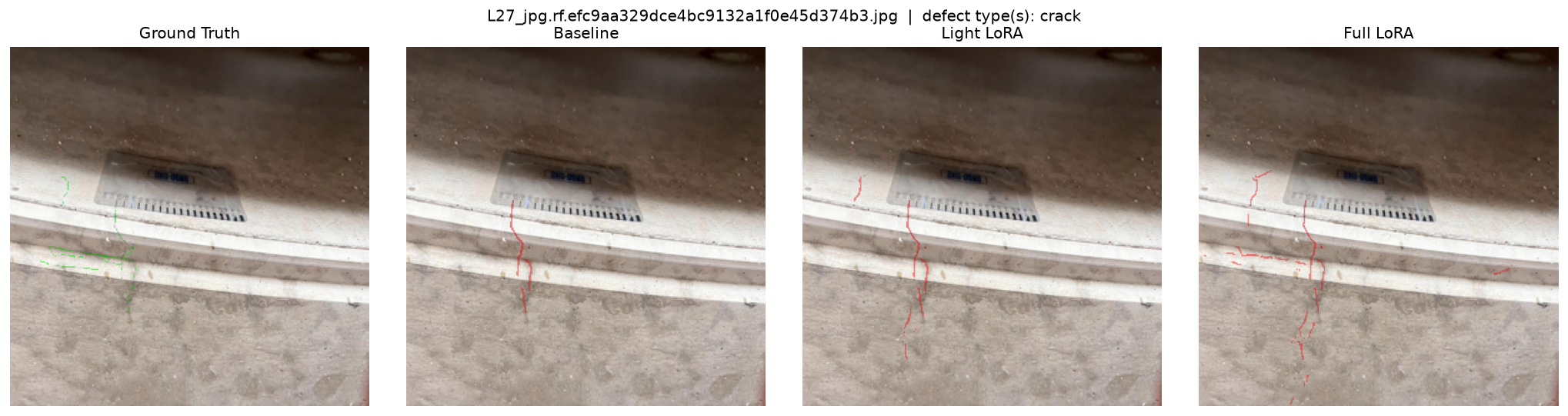}
\caption{Tunnel test image L27 (crack). Full recovers more of the true branching structure than Light or the baseline, illustrating that the aggregate preference for Light does not hold in every individual case.}
\label{fig:10}
\end{figure*}

\begin{figure*}[!t]
\centering
\includegraphics[width=\textwidth]{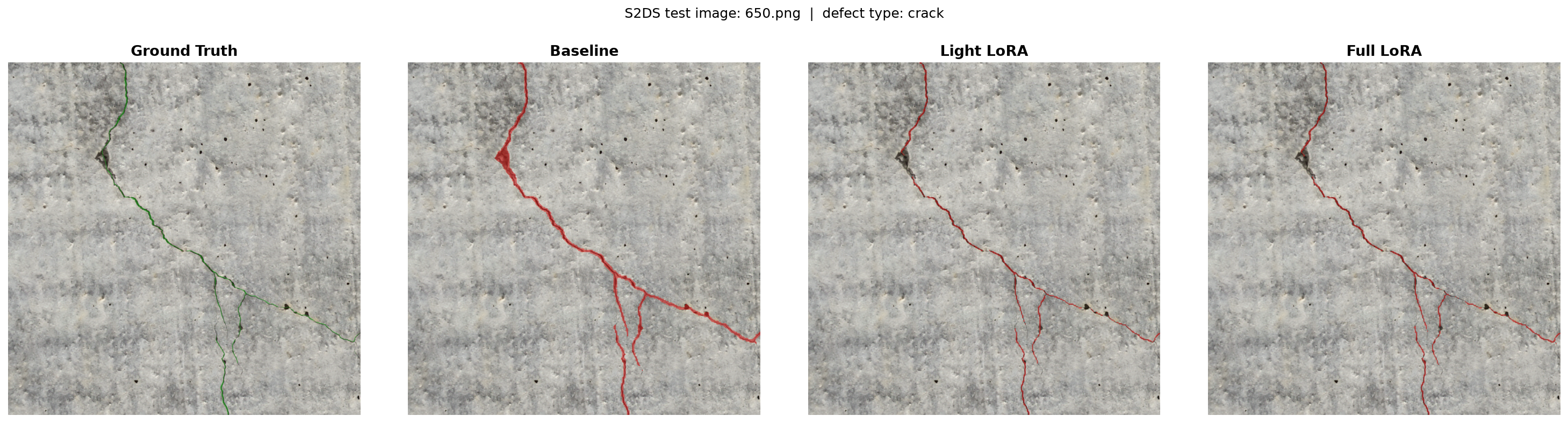}
\caption{S2DS test image 650 (crack). All three models track the branching crack; the baseline shows a distinctly thicker, less precise patch at the point of origin.}
\label{fig:11}
\end{figure*}

\begin{figure*}[!t]
\centering
\includegraphics[width=\textwidth]{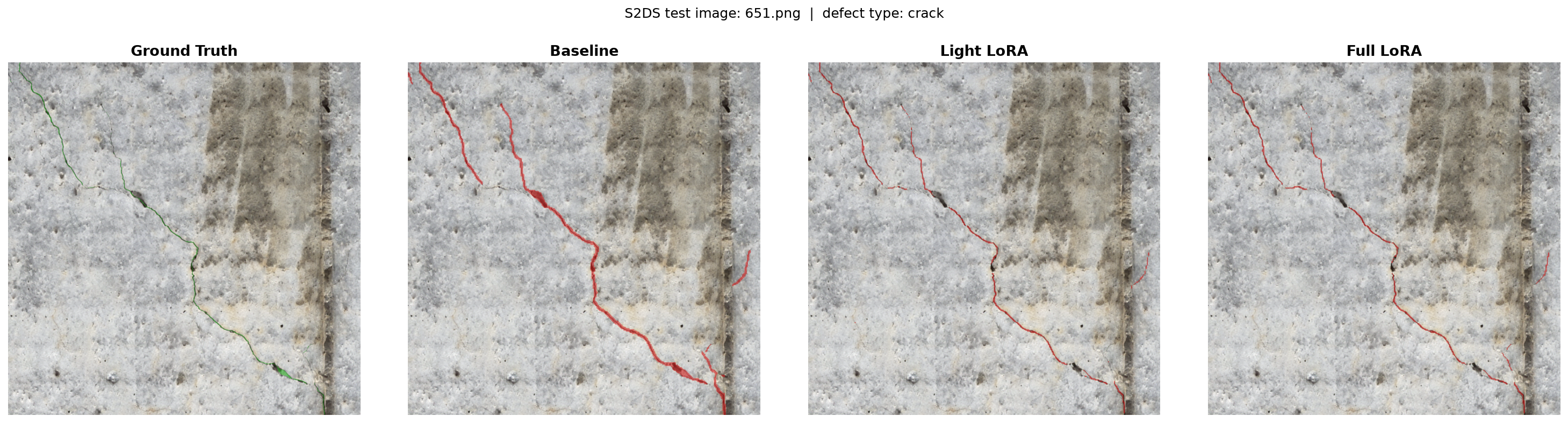}
\caption{S2DS test image 651 (crack). All three perform comparably on the diagonal crack pattern, with only minor differences.}
\label{fig:12}
\end{figure*}

\begin{figure*}[!t]
\centering
\includegraphics[width=\textwidth]{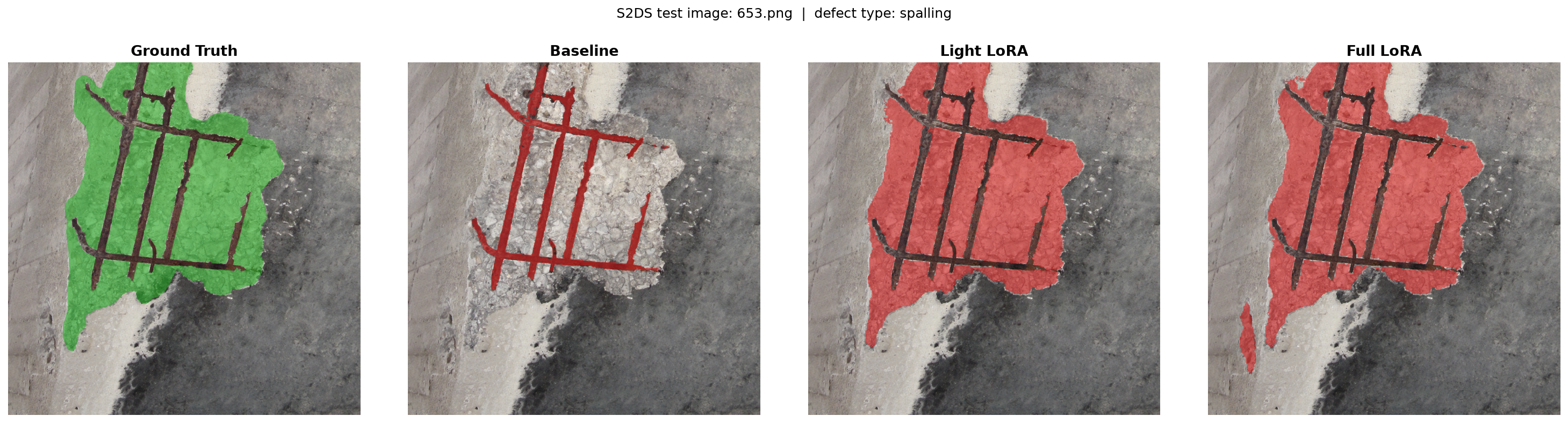}
\caption{S2DS test image 653 (spalling). The baseline's raw-confidence prediction approximates the irregular boundary but falls below the presence gate and is scored as zero recall; Light and Full both match the true region closely.}
\label{fig:13}
\end{figure*}

\begin{figure*}[!t]
\centering
\includegraphics[width=\textwidth]{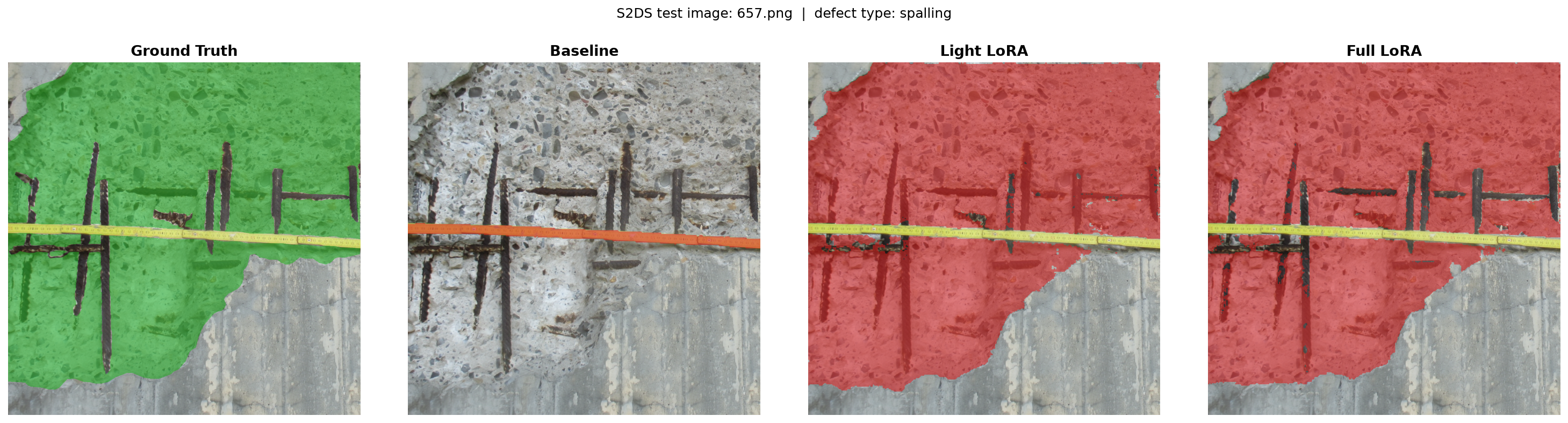}
\caption{S2DS test image 657 (spalling). The baseline produces no visible detection; Light and Full achieve near-total coverage of the irregular region despite the embedded measuring scale and rebar.}
\label{fig:14}
\end{figure*}

Figs.~\ref{fig:3}--\ref{fig:10} present representative tunnel test predictions and Figs.~\ref{fig:11}--\ref{fig:14} representative S2DS predictions, each in a four-panel layout (ground truth, baseline, Light, Full). Panels were selected to illustrate specific behaviors rather than sampled at random, and should be read as such.

Two failure modes of the frozen baseline are visible. The first is large-area over-prediction, in which nearly the entire frame is returned as defect (Figs.~\ref{fig:3} and~\ref{fig:4}); both adapted configurations localize the true regions instead. The second is silence on categories outside the pretraining distribution, with only a faint low-confidence trace on water ingress (Figs.~\ref{fig:5} and~\ref{fig:6}) or no detection at all on S2DS spalling (Fig.~\ref{fig:14}). Fig.~\ref{fig:13} shows an intermediate case in which the baseline's raw-confidence prediction approximates the spalling boundary but falls below the presence gate and is scored as zero recall --- a reminder that presence gating, not mask quality alone, determines the reported baseline numbers on that category.

Among adapted configurations the differences are smaller and not uniformly in one direction. Light produces the more contiguous, better-bounded spalling predictions in Figs.~\ref{fig:7} and~\ref{fig:8}, where Full is fragmented in one case and adds a spurious region in the other; Full recovers more of a branching crack structure in Fig.~\ref{fig:10}. This mirrors the aggregate finding that neither configuration dominates.

\section{Discussion}\label{discussion}

\subsection{The efficiency gain is a property of the method rather than of one dataset}\label{the-efficiency-gain-is-a-property-of-the-method-rather-than-of-one-dataset}

Pixel IoU improved approximately 20-fold on the tunnel dataset and 52-fold on S2DS, and instance-level recall by a closely matched 1.7- to 1.8-fold on both. Consistency of direction across ten independently computed metrics on two independently collected datasets, under an identical protocol, supports attributing the gain to the adaptation method rather than to properties of a single dataset's imagery or annotation conventions.

The magnitude of the gain is plausibly explained by what the frozen model lacks rather than by refinement of what it has. On both datasets the baseline scored exactly zero pixel IoU on the category with no natural-image analogue --- water ingress and spalling respectively --- which is consistent with adaptation constructing task-relevant representations from a small in-domain training set rather than sharpening an existing but weak one. Section~\ref{benefit-is-concentrated-where-zero-shot-competence-is-absent} develops this reading from the per-category results.

The scope of the claim is bounded by the evidence: two datasets, both concrete and tunnel-lining surface defects. Transfer to visually distinct inspection domains such as steel or timber remains untested, and a third structurally distinct benchmark would be required to establish the boundary.

\subsection{Adapter capacity, and a confound}\label{adapter-capacity-and-a-confound}

Neither placement dominated. On the tunnel dataset Full favored detection-oriented metrics and Light pixel-level generalization; on S2DS the two were close, with a small consistent pixel-IoU edge to Light. A plausible mechanism is visible in the training dynamics: Full's larger footprint fits the training distribution faster, as reflected in its consistently earlier best checkpoint, which can improve detection metrics before overfitting onset without conferring an equivalent pixel-level generalization benefit.

This interpretation must be qualified. The tunnel and S2DS experiments used different implementation revisions, so the configurations sharing the label ``Light'' adapt materially different module sets (124 attention projections across fusion encoder, decoder, and head, versus 67 vision-backbone MLP layers plus one attention module). The within-dataset comparisons remain valid, but the cross-dataset capacity comparison cannot be separated from this implementation difference, and we therefore do not draw capacity-selection guidance from it. Re-running the S2DS comparison under the tunnel study's adapter definitions is the necessary next step, and is a prerequisite to any general claim about where adapters should be placed.

Within the tunnel dataset, where the comparison is controlled, the result of practical interest is that an 11.2-fold increase in trainable parameters did not produce a uniform improvement: the smaller placement achieved the higher pixel IoU. This is consistent with the view that adapter \emph{placement} interacts with the target task at least as strongly as adapter \emph{capacity}, and argues against defaulting to the largest footprint that memory permits.

\subsection{Benefit is concentrated where zero-shot competence is absent}\label{benefit-is-concentrated-where-zero-shot-competence-is-absent}

The per-category results show a consistent relationship on both datasets: the categories the frozen model handles least well gain most from adaptation, and the category it partially handles gains least. Crack, the majority training category in both datasets at roughly 71\% of annotations, attained the lowest post-adaptation pixel IoU in both (0.337 tunnel, 0.4317 S2DS), improving 1.7-fold and 2.8-fold. Spalling and water ingress, starting from baselines of 0.011 and 0.000 on the tunnel dataset and 0.0000 on S2DS, reached comparable or higher absolute values.

Two implications follow. First, training-set frequency is a poor predictor of per-category benefit: the majority class benefited least on both datasets. Second, the value of adapting a foundation model is greatest precisely for the categories a general-purpose pretraining distribution does not cover, which is an argument for parameter-efficient adaptation on exactly the specialized categories that make a domain difficult, rather than on the categories that are already well represented.

We note that this reading revises an interpretation that a single-dataset analysis would support. Considering the tunnel dataset alone, crack has the strongest baseline and might be described as the category the method suits best; considering post-adaptation values alone, it is the weakest. Only the paired comparison of baseline and adapted values, across two datasets, makes the underlying relationship legible. The finding is limited to the two categories shared between the datasets, since water ingress appears in only one.

\subsection{Thin-structure geometry and the choice of metric}\label{thin-structure-geometry-and-the-choice-of-metric}

Crack's low pixel IoU coexists with high pixel recall in both datasets, and in the frozen baseline the combination is extreme (recall 0.9911 with IoU 0.1533 on S2DS). This is the expected signature of a thin elongated target. For a structure a few pixels wide, a boundary error of one or two pixels changes the union substantially while barely changing the intersection, so IoU is bounded well below what visually accurate segmentation would suggest. The instance-matching metrics compound this through the fragmentation penalty described in Section~\ref{evaluation-metrics}, in which a single crack predicted as two disjoint portions is scored as one missed instance and two false positives.

The consequence for evaluation practice is that absolute mAP values for crack-dominated datasets are not comparable to those for datasets of compact objects, and that reporting instance metrics alone would misrepresent the adapted models --- the gap between pixel IoU (0.8554) and mAP@50 (0.2207) on S2DS is a measurement property of the target geometry as much as a property of the model. This does not excuse the low absolute detection numbers, which remain a genuine limitation, but it does mean that closing them is partly a question of prediction connectivity rather than of localization accuracy, and that fragment-merging post-processing is the natural first intervention.

\subsection{Deployment implications and residual domain gap}\label{deployment-implications-and-residual-domain-gap}

Both configurations train fewer than 1.4\% of model parameters, a reduction exceeding 98.6\% relative to full fine-tuning, and both were trained on a single workstation rather than a cluster. Combined with adapter-only checkpoints of tens of megabytes, this makes maintaining several site-specific adaptations of one frozen base model practical for organizations without dedicated infrastructure --- which is the deployment setting that motivated the study.

The absolute results nevertheless differ substantially between domains. The best test pixel IoU was 0.338 on the tunnel dataset against 0.8554 on S2DS, a gap of roughly 2.5-fold, despite near-identical frozen baselines (0.017 and 0.0165). The tunnel dataset presents frequent multi-defect co-occurrence within a single frame and a curved, segmented lining geometry with bolts and joints, against which S2DS's flatter, single-defect concrete surfaces are a simpler scene. This attribution is unverified: no controlled ablation isolates scene complexity from other differences such as annotation density, augmentation, and image resolution, and the datasets also differ in the number of categories. Multi-instance-aware prompting and geometry-conditioned augmentation are the directions we consider most likely to narrow the gap.

\subsection{Limitations}\label{limitations}

The principal limitations are stated here together. (i) Only two datasets were evaluated, both concrete and tunnel-lining surface defects, so generalization beyond this visual domain is untested. (ii) The cross-dataset adapter-capacity comparison is confounded by differing implementation revisions (Section~\ref{adapter-capacity-and-a-confound}). (iii) Absolute detection metrics remain low on both datasets, partly for the geometric reason analyzed in Section~\ref{thin-structure-geometry-and-the-choice-of-metric} and partly as a genuine limitation. (iv) The presence--text decoupling of Section~\ref{hard-negative-prompt-supervision} was diagnosed and mitigated, but the study reports no quantitative negative-prompt benchmark; establishing a metric for prompt discrimination, and ablating the two negative tiers against it, is necessary to convert the observation into a measured result. (v) Per-category evaluation units are few in some cells (as low as \(n = 5\) for tunnel water ingress), so per-category values should be read as indicative rather than precise.

\section{Conclusion}\label{conclusion}

This study examined Low-Rank Adaptation of SAM3, a concept-promptable segmentation foundation model, for multi-class structural defect segmentation, and addressed both how such a model can be supervised from conventional class-labeled annotation and whether the resulting efficiency gain transfers across datasets.

Two methodological results are the primary contribution. A concept-promptable model can be trained directly from COCO-style instance segmentation by using category names verbatim as prompts and grouping annotations into exhaustive per-concept queries, without prompt templates or learned class embeddings, provided the label set is linguistically meaningful. Doing so, however, supplies positive prompts exclusively, which causes the model's presence estimate to decouple from the text condition and respond to any prompt --- a collapse invisible to conventional positive-only metrics. Exhaustive hard-negative prompting, issuing every absent dataset category as a zero-detection query, mitigates this at no annotation cost, and generalizes beyond the specific negatives used because the text encoder remains frozen.

Empirically, two adapter placements updating 0.121\% and 1.341\% of parameters substantially outperformed the frozen baseline on both a purpose-built tunnel lining dataset and the independent public S2DS benchmark, with pixel IoU improving approximately 20-fold and 52-fold and instance-level recall by a consistent 1.7- to 1.8-fold. Neither placement dominated, and the larger footprint did not yield uniformly better results, indicating that adapter placement interacts with the task at least as strongly as capacity. Per-category analysis showed that benefit concentrates where zero-shot competence is absent rather than where training data is abundant: the majority category gained least on both datasets.

Taken together, these results indicate that parameter-efficient adaptation offers a practical route to specialized deployment of promptable segmentation foundation models on a single consumer-grade workstation, and that the supervision design --- how classes become prompts, and what the model is taught \emph{not} to respond to --- deserves the same attention as the choice of adapter.

\section{Code and Data Availability}\label{code-and-data-availability}

The SAM3-LoRA implementation used in this study is publicly available at https://github.com/Sompote/sam3\_lora, developed by the AI Research Group, KMUTT (2026). S2DS is publicly available from Benz and Rodehorst \cite{ref9}.

\section{Declaration of Competing Interest}\label{declaration-of-competing-interest}

The authors declare that they have no known competing financial interests or personal relationships that could have appeared to influence the work reported in this paper.

\section{Declaration of Generative AI and AI-assisted Technologies in the Writing Process}\label{declaration-of-generative-ai-and-ai-assisted-technologies-in-the-writing-process}

During the preparation of this work the authors used Anthropic's Claude and Google's Gemini to improve language and readability. After using these tools, the authors reviewed and edited the content as needed and take full responsibility for the content of the publication.

\section{Acknowledgments}\label{acknowledgments}

This research received funding support from the National Science and Technology Development Agency (NSTDA) through the National Science, Research and Innovation Fund (NSRF) via the Program Management Unit for Human Resources \& Institutional Development, Research and Innovation (grant number B13F670082-11). The authors thank the construction companies and infrastructure inspection agencies that provided access to field data collection sites, and the domain experts who assisted with data annotation and validation. Computational resources provided by the AI Research Group at King Mongkut's University of Technology Thonburi are gratefully acknowledged.

\balance

\end{document}